\documentclass[wcp]{jmlr}

\usepackage{booktabs}
\usepackage{multirow}
\usepackage{tikz}
\usetikzlibrary{arrows.meta,positioning,fit,backgrounds,calc}
\usepackage{xcolor}
\usepackage{colortbl}
\usepackage{capt-of}

\graphicspath{{figures/}}

\AtBeginDocument{%
  \setcounter{topnumber}{4}%
  \setcounter{bottomnumber}{3}%
  \setcounter{totalnumber}{6}%
  \setlength{\textfloatsep}{10pt plus 2pt minus 2pt}%
  \setlength{\floatsep}{8pt plus 2pt minus 2pt}%
  \setlength{\intextsep}{8pt plus 2pt minus 2pt}%
  \setlength{\abovecaptionskip}{5pt}%
  \setlength{\belowcaptionskip}{0pt}%
}
\makeatletter
\renewcommand{\section}{\@startsection{section}{1}{\z@}%
   {-0.18in \@plus -1ex \@minus -.2ex}%
   {0.08in \@plus.2ex}%
   {\normalfont\rmfamily\bfseries\large\raggedright}}
\renewcommand\subsection{\@startsection{subsection}{2}{\z@}%
   {-0.15in \@plus -1ex \@minus -.2ex}%
   {0.06in \@plus .2ex}%
   {\normalfont\rmfamily\bfseries\normalsize\raggedright}}
\renewcommand\paragraph{\@startsection{paragraph}{4}{\z@}%
   {1.2ex plus 0.5ex minus .2ex}%
   {-1em}%
   {\normalfont\normalsize\rmfamily\bfseries}}
\g@addto@macro\normalsize{%
  \setlength\abovedisplayskip{7pt plus 2pt minus 3pt}%
  \setlength\belowdisplayskip{7pt plus 2pt minus 3pt}%
  \setlength\abovedisplayshortskip{1pt plus 2pt}%
  \setlength\belowdisplayshortskip{4pt plus 2pt minus 2pt}}
\makeatother

\newcommand{\dsvd}{Delta-SVD}
\newcommand{\dW}{\Delta W}
\newcommand{\Er}{E_{r}}
\definecolor{dsvdblue}{HTML}{2a78d6}
\definecolor{dsvdred}{HTML}{e34948}
\definecolor{gpa}{RGB}{30,90,160}
\definecolor{gpb}{RGB}{160,40,40}

\jmlrvolume{353,}
\jmlryear{2026}
\jmlrworkshop{ACML 2026}

\usepackage{lineno}

\title[Delta-SVD: Efficient Compression for Personalized Text-to-Image Models]{Delta-SVD: Efficient Compression for Personalized Text-to-Image Models}

\author[Zhang Chen Chen Yan Ye Cai]{%
\Name{Tangyuan Zhang\nametag{$^{1,2}$}} \Email{Tangyuan.Zhang@student.rmit.edu.au}\\
\Name{Shangyu Chen\nametag{$^{1}$}} \Email{Shangyu.Chen@monash.edu}\\
\Name{Qixiang Chen\nametag{$^{1}$}} \Email{Qixiang.Chen@monash.edu}\\
\Name{Junyu Yan\nametag{$^{1}$}} \Email{jyan0349@student.monash.edu}\\
\Name{Jingwen Ye\nametag{$^{1}$}} \Email{Jingwen.Ye@monash.edu}\\
\Name{Jianfei Cai\nametag{$^{1}$}} \Email{Jianfei.Cai@monash.edu}\\
\addr $^{1}$Department of Data Science and AI, Monash University, Australia\\
\addr $^{2}$School of Computing Technologies, RMIT University, Australia}

\editors{Andy Song, Bo Han and Sarah Erfani}

\hypersetup{hypertexnames=false,pdftitle={Delta-SVD: Efficient Compression for Personalized Text-to-Image Models},pdfauthor={Tangyuan Zhang, Shangyu Chen, Qixiang Chen, Junyu Yan, Jingwen Ye, Jianfei Cai}}

\begin{document}

\makeatletter
\let \@jmlrpages \@empty
\makeatother

\maketitle

\begin{abstract}
Personalizing a text-to-image diffusion model by full fine-tuning gives the strongest subject fidelity but produces a full copy of the model per subject: a thousand personalized Stable Diffusion checkpoints approach five terabytes, so storage, not compute, is what fails to scale. We present \textbf{\dsvd{}}, which compresses the \emph{weight change} that personalization makes rather than the model. Given only a finished checkpoint and its base, it takes a truncated SVD of each layer's weight delta and keeps the few components that carry its spectral energy, selected by a per-layer energy threshold or by a global byte budget filled by water-filling, optimal for the continuous relaxation and empirically near-optimal for discrete rank allocation. Nothing is trained and no subject image is read, so the method applies to legacy and third-party checkpoints for which training-time adapters such as LoRA were never produced. The factors merge back into dense weights at load time, so serving runs at the uncompressed model's exact latency, whereas whole-model 8-bit and 4-bit quantization slow inference by $1.5$--$3.4\times$. On the $30$-subject DreamBooth benchmark, \dsvd{} matches the uncompressed model to within $0.01$ DINO at $7\times$ compression and keeps $98\%$ of its fidelity at $33\times$. Among post-hoc baselines, only dense-delta quantization at $411$\,MiB matches it; below that size \dsvd{} dominates pruning, random-rank selection, and both whole-model and dense-delta quantization on the storage--quality frontier. The same holds on $84$ diverse CustomConcept101 subjects, where a $9$\,MiB adapter reproduces the uncompressed outputs at paired DINO $0.77$, and on a fully fine-tuned SD3.5-medium MMDiT, where a $531$\,MB adapter matches the uncompressed model to within $0.01$ DINO.
\end{abstract}


\section{Introduction}
\label{sec:intro}

Text-to-image diffusion models~\citep{ho2020ddpm,rombach2022ldm} are now routinely personalized: a few photos of a pet, a product, or a person yield a model that renders that subject in new scenes. The recipes with the highest fidelity update the entire denoising network~\citep{ruiz2023dreambooth}, so each subject is a full copy of the model, about $5$\,GB for Stable Diffusion v1.5. Training is paid once and inference costs no more than the base model; storage grows with the catalog, and a thousand subjects approach $5$\,TB.

The catalogs that need compressing are rarely ones we could have built differently. Their checkpoints are inherited, trained by another party or downloaded from a hub, and the subject photos behind them are often gone: deleted under a retention policy, locked by licensing, or withheld for privacy. Compression must therefore be \emph{post-hoc}, \emph{training-free}, and \emph{data-free}. Most efficient-personalization work does not fit: LoRA~\citep{hu2022lora} and SVDiff~\citep{han2023svdiff} fix a compact parameterization \emph{before} fine-tuning and cannot be applied to a full checkpoint afterwards, and distillation~\citep{kim2024bksdm} needs data and training. Post-training quantization and pruning do fit, but neither covers the whole size range: whole-model quantization compresses by less than $2\times$ and slows inference, pruning the weight change destroys identity, and quantizing it is lossless down to about $400$\,MiB but offers nothing useful below (Section~\ref{sec:exp}).

The right object to compress is the change personalization made to the model: all subject-specific information lies in the delta $\dW = W_{\text{ft}} - W_0$, and this delta is strongly low-rank: on the SD-v1.5 UNet, $90\%$ of each layer's spectral energy lies in $19\%$ of its singular components, and on the SD3.5 MMDiT in under $8\%$ (Section~\ref{sec:lowrank}). The truncated SVD gives the best low-rank approximation of that change in closed form, with no learning.

\begin{figure}[t]
\centering
\includegraphics[width=0.93\textwidth,trim=8 8 11 9,clip]{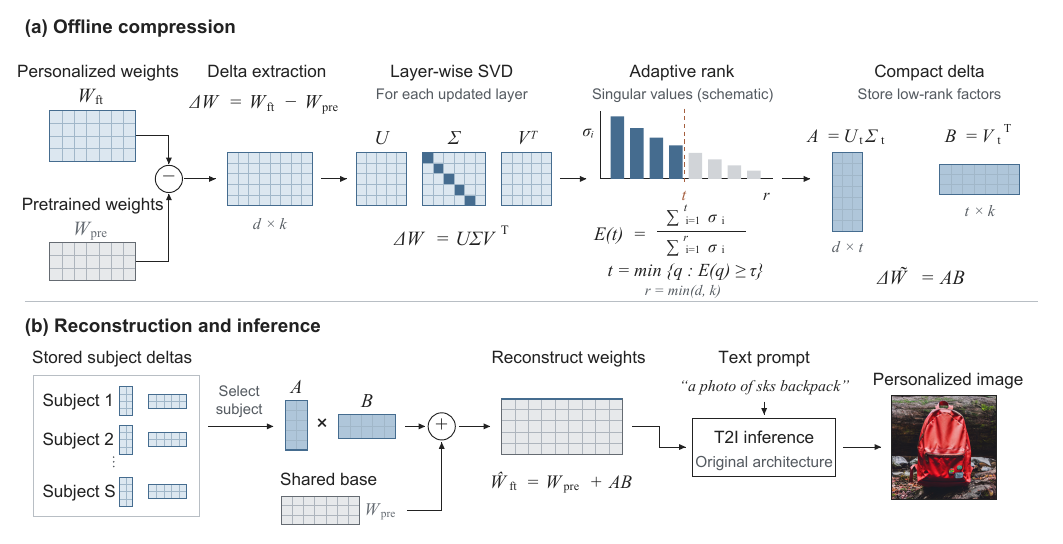}
\caption{\textbf{\dsvd{} overview.} (a) Each layer's delta $\dW = W_{\text{ft}} - W_{\text{pre}}$ ($W_0$ in the text) is truncated by SVD to the components carrying its spectral energy, selected by a threshold $\tau$ or a byte budget; the factors $A = U_t\Sigma_t$, $B = V_t^{\top}$ are the only per-subject artifact. (b) At load time $\widetilde{\dW} = AB$ is merged into the shared base, so serving uses a standard dense checkpoint; no training and no subject images are needed.}
\label{fig:pipeline}
\end{figure}

\textbf{\dsvd{}} (Figure~\ref{fig:pipeline}) stores the leading SVD factors of each layer's delta as a compact adapter; at load time they are multiplied out and added back onto the frozen base, so the served network is an ordinary dense model with the latency and memory of the uncompressed one. Each layer's rank is set by a spectral-energy threshold $\tau$ or by a global byte budget that a water-filling rule distributes across layers, optimally for its continuous relaxation and near-optimally in practice (Section~\ref{sec:method}). Compression takes seconds to minutes. On the DreamBooth benchmark, a $7\times$ smaller adapter matches the uncompressed model to within $0.01$ DINO, and a $33\times$ smaller one keeps $98\%$ of its subject fidelity.

\paragraph{Contributions.}
\begin{enumerate}\setlength{\itemsep}{1pt}
\item \textbf{Low-rank personalization deltas and a closed-form compressor.} We characterize the spectra of full fine-tuning deltas on a UNet and an MMDiT (Section~\ref{sec:lowrank}) and build a post-hoc, training-free, data-free compressor with two rank-selection rules, per-layer energy truncation and byte-budget water-filling (Section~\ref{sec:method}).
\item \textbf{A controlled storage--quality study} on the DreamBooth benchmark (Section~\ref{sec:exp}). Below about $400$\,MiB, \dsvd{} dominates every post-hoc baseline we test (whole-model and dense-delta quantization, magnitude pruning, random-rank selection) at no added inference cost. Adapters trained from scratch with the data can be competitive at matched size; \dsvd{}'s advantage is fidelity to the \emph{existing checkpoint} and its outputs, which a catalog of inherited models requires.
\item \textbf{Generality} (Section~\ref{sec:general}). The same trade-offs hold on $84$ CustomConcept101 subjects, and on a fully fine-tuned SD3.5-medium MMDiT, where a $531$\,MB adapter matches the uncompressed model to within $0.01$ DINO. Merged-LoRA checkpoints on SDXL and FLUX.1 are reported separately, since their rank cap makes them weaker evidence.
\item \textbf{A deployment analysis} (Section~\ref{sec:deploy}). \dsvd{} serves at the uncompressed model's latency, whereas whole-model quantization slows inference by $1.5$--$3.4\times$; we measure load time in two serving regimes and catalog storage, where a thousand subjects shrink from $4.8$\,TiB to $0.14$\,TiB at the quality knee.
\end{enumerate}

\section{Related Work}
\label{sec:related}

\paragraph{Personalization and training-time adapters.}
DreamBooth~\citep{ruiz2023dreambooth} fine-tunes the whole denoising network on a few subject images and gives the strongest identity fidelity; it is the recipe we compress. Cheaper alternatives fix a compact parameterization before training: Textual Inversion~\citep{gal2023textualinversion} learns a token embedding, Custom Diffusion~\citep{kumari2023customdiffusion} updates cross-attention keys and values and a token embedding, LoRA~\citep{hu2022lora} trains a low-rank update, and SVDiff~\citep{han2023svdiff} learns a shift of the base weights' singular values. AdaLoRA~\citep{zhang2023adalora} and SARA~\citep{gu2024sara} reallocate rank from an SVD, PiSSA~\citep{meng2024pissa} initializes from the base weights' top singular components, and TriLoRA~\citep{feng2024trilora} and LyCORIS~\citep{yeh2024lycoris} change the adapter's factorization, all during fine-tuning; our allocator (Section~\ref{sec:budget}) works after it. All need data and a training run. SVDiff in particular decomposes the pretrained weights $W_0 = U_0 \Sigma_0 V_0^{\top}$, whereas \dsvd{} decomposes the weight change $\dW$ after fine-tuning: SVDiff is a way to train, \dsvd{} a way to store (Section~\ref{sec:exp} compares them).

\paragraph{Post-hoc compression of diffusion models.}
Post-training quantization~\citep{li2023qdiffusion,he2023ptqd}, low-rank-assisted 4-bit inference~\citep{li2024svdquant}, structural pruning~\citep{fang2023diffpruning}, and distillation~\citep{kim2024bksdm} shrink the shared base and the cost of running one model; \dsvd{} shrinks the cost of storing each personalized variant on top of it, so the two are orthogonal.

\paragraph{Delta compression in LLM serving.}
The same problem arises when many full fine-tunes of one language model are served together. DeltaZip~\citep{yao2025deltazip} sparsifies and quantizes deltas in a co-designed stack, BitDelta~\citep{liu2024bitdelta} keeps signs plus distilled per-matrix scales, Delta-CoMe~\citep{ping2024deltacome} quantizes the delta's singular vectors at mixed precision, and PHLoRA~\citep{vasani2025phlora}, the closest to ours, extracts a low-rank adapter from a full fine-tune without data. The first three target language-model accuracy, allocate bits over a fixed decomposition rather than rank under a byte objective, and need custom kernels rather than a dense merge; PHLoRA merges densely but scores task accuracy, and relative to it we add a byte-budget allocator with a proof of optimality for its continuous relaxation, three subject sets, two fully fine-tuned architectures, and evaluation by perceptual identity.

\section{Personalization Deltas are Low-Rank}
\label{sec:lowrank}

\begin{figure}[t]
\centering
\begin{minipage}[c]{0.40\textwidth}
\centering
\includegraphics[width=\linewidth]{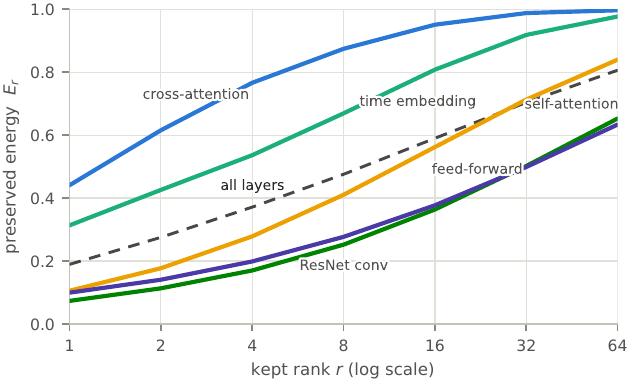}
\caption{\textbf{Preserved $\sigma^2$-energy of $\dW$ versus kept rank}, SD-v1.5 UNet, mean over $30$ subjects by layer family. Cross-attention concentrates fastest; feed-forward and ResNet convolutions are flattest.}
\label{fig:energy}
\end{minipage}\hfill
\begin{minipage}[c]{0.57\textwidth}
\centering
\captionof{table}{\textbf{Effective rank of the personalization delta on two fully fine-tuned backbones.} rank$_{e90}$ (\%) is the fraction of singular components holding $90\%$ of the $\sigma^2$-energy and is comparable across architectures; stable rank is $\|\dW\|_F^2/\|\dW\|_2^2$, mean $\pm$ std over subjects; full rank is the typical per-layer full rank.}
\label{tab:rankstats}
\small
\setlength{\tabcolsep}{3pt}
\resizebox{\linewidth}{!}{\begin{tabular}{llccccc}
\toprule
Backbone & Arch. & \#subj. & full rank & stable rank & rank$_{e90}$ (\%) & rank$_{e95}$ (\%) \\
\midrule
SD-v1.5       & UNet  & 30 & 570  & $7.94 \pm 0.53$ & 19.0 & 28.3 \\
SD3.5-medium  & MMDiT & 12 & 1256 & $5.36 \pm 0.29$ & \phantom{0}7.9 & 14.5 \\
\bottomrule
\end{tabular}}
\end{minipage}
\end{figure}

For every personalized checkpoint we form the per-layer delta $\dW^{(l)} = W^{(l)}_{\text{ft}} - W^{(l)}_0$, flatten convolution kernels into matrices, and compute the full singular spectrum of each of the $282$ weight matrices in the SD-v1.5 UNet (the $109$ normalization vectors are stored as they are). Figure~\ref{fig:energy} reports the $\sigma^2$-energy preserved at a given rank; Table~\ref{tab:rankstats} the fraction of the spectrum needed for a given energy and the stable rank $\|\dW\|_F^2/\|\dW\|_2^2$.

The spectrum concentrates quickly: averaged over $30$ subjects, $90\%$ of a layer's energy lies in the top $19\%$ of its singular components, and the stable rank is about $8$ against a typical full rank of $570$. It is also uneven across layer families: cross-attention deltas concentrate fastest, time-embedding and attention projections follow, and feed-forward and ResNet convolutions are flattest. This agrees with Custom Diffusion~\citep{kumari2023customdiffusion}, where updating cross-attention alone captures a subject, here in a model free to update everything.

The same holds on a fully fine-tuned transformer. On the $2.2$B SD3.5-medium MMDiT ($12$ subjects, same recipe) the stable rank is $5.4$ and $90\%$ of the energy lies in $7.9\%$ of the spectrum, against $19\%$ for the UNet; we read no scale trend into this pair, since the backbones differ in architecture as well as size. Merged-LoRA deltas (SDXL, FLUX.1) are rank-bounded by construction and cannot measure the intrinsic rank of full fine-tuning; Section~\ref{sec:merged} reports them separately.

\section{\dsvd{}}
\label{sec:method}

\subsection{Setting}
\label{sec:problem}
Let $W_0 = \{W_0^{(l)}\}_{l=1}^{L}$ be the weights of a frozen base denoising network and $W_{\text{ft}}$ those of a checkpoint personalized to one subject by full fine-tuning. Given only these weights, with no subject images and no training, we seek an adapter $\mathcal{A}$ with $\lvert \mathcal{A}\rvert \ll \lvert W_{\text{ft}}\rvert$ whose merge into $W_0$ reproduces the behavior of $W_{\text{ft}}$, the regime of legacy and third-party checkpoints.

\subsection{Truncating the delta}
\label{sec:deltasvd}
For each layer we form
\begin{equation}
\dW^{(l)} \;=\; W_{\text{ft}}^{(l)} - W_0^{(l)} \;\in\; \mathbb{R}^{m_l \times n_l},
\label{eq:delta}
\end{equation}
merging the spatial and input-channel axes of convolutional kernels into the column dimension. With the singular value decomposition
\begin{equation}
\dW^{(l)} \;=\; U^{(l)} \Sigma^{(l)} {V^{(l)}}^{\top}, \qquad
\Sigma^{(l)} = \mathrm{diag}\!\big(\sigma_1^{(l)} \ge \cdots \ge \sigma_{\rho_l}^{(l)}\big),\; \rho_l = \min(m_l,n_l),
\label{eq:svd}
\end{equation}
the rank-$r$ truncation
\begin{equation}
\widehat{\dW}^{(l)}_{r} \;=\; \sum_{i=1}^{r} \sigma_i^{(l)}\, u_i^{(l)} {v_i^{(l)}}^{\top}
\label{eq:trunc}
\end{equation}
is the Frobenius-optimal rank-$r$ approximation~\citep{eckart1936approximation}, and it preserves the energy fraction
\begin{equation}
\Er^{(l)} \;=\; \frac{\sum_{i=1}^{r} \big(\sigma_i^{(l)}\big)^2}{\sum_{i=1}^{\rho_l} \big(\sigma_i^{(l)}\big)^2}.
\label{eq:energy}
\end{equation}
The adapter stores $\big(U^{(l)}_{:,:r}\Sigma^{(l)}_{:r},\, V^{(l)}_{:,:r}\big)$: $r(m_l+n_l)$ numbers in place of $m_l n_l$. At load time the factors are multiplied out once and merged,
\begin{equation}
\widehat{W}^{(l)} \;=\; W_0^{(l)} + \alpha\,\widehat{\dW}^{(l)}_{r},
\label{eq:merge}
\end{equation}
with $\alpha{=}1$ unless stated. The result is a dense weight of the original shape and dtype, so the served network has the same graph, latency, and peak memory as the uncompressed model.

\subsection{Energy-threshold selection}
\label{sec:energytau}
The simpler rule fixes one scalar $\tau \in (0,1]$ and gives each layer the smallest rank whose cumulative spectrum reaches $\tau$:
\begin{equation}
r_l(\tau) \;=\; \min\Big\{ r : \textstyle\sum_{i=1}^{r} \sigma_i^{(l)} \,\big/\, \sum_{i=1}^{\rho_l} \sigma_i^{(l)} \;\ge\; \tau \Big\}.
\label{eq:tauselect}
\end{equation}
Layers whose delta concentrates faster keep fewer components, and sweeping $\tau$ traces the storage--quality curve of Section~\ref{sec:exp}.\footnote{Rule \eqref{eq:tauselect} thresholds the cumulative sum of $\sigma$, as all reported $\tau$ checkpoints were produced, whereas $\Er$ in \eqref{eq:energy} uses $\sigma^2$; the two are monotonically related, and on SD-v1.5 $\tau{=}0.2$ and $0.5$ correspond to $0.68$ and $0.90$ preserved $\sigma^2$-energy.}

\subsection{Byte-budget allocation by water-filling}
\label{sec:budget}
A single $\tau$ ignores that layers differ in cost: reaching it in a slowly concentrating layer may gain little energy for many bytes. For a hard storage target we solve
\begin{equation}
\max_{\{r_l\}} \; \sum_{l=1}^{L} \sum_{i=1}^{r_l} \big(\sigma_i^{(l)}\big)^2
\quad \text{s.t.} \quad
\sum_{l=1}^{L} r_l\, c_l \;\le\; B, \qquad c_l = (m_l + n_l)\, b,
\label{eq:budget}
\end{equation}
where $b$ is bytes per stored parameter, $c_l$ the cost of one rank increment in layer $l$, and $B$ the factor-payload budget net of fixed serialization overhead. Write $v_{li} = (\sigma_i^{(l)})^2$, non-increasing in $i$, and let $x_{li}\in[0,1]$ be the kept fraction of direction $i$ of layer $l$, with prefix constraint $x_{l1}\ge x_{l2}\ge\cdots$. Water-filling sorts all directions by energy per byte $v_{li}/c_l$ and fills the budget in that order.

\begin{proposition}[Optimality of water-filling]
\label{prop:waterfill}
Filling the budget in decreasing order of $v_{li}/c_l$, allowing a fraction of the final direction, maximizes retained energy over the continuous relaxation $x_{li}\in[0,1]$ of \eqref{eq:budget}. The resulting allocation is also optimal for the integer problem $x_{li}\in\{0,1\}$ whenever it is integral, and whenever all layers share one cost $c_l = c$, in which case it reduces to keeping the $\lfloor B/c\rfloor$ largest energies.
\end{proposition}

Appendix~\ref{app:proof} gives the Lagrangian proof and a penalized-objective corollary. The implemented allocator keeps whole directions and runs as one heap pass over the sorted spectra in under a second; with heterogeneous UNet layer costs it can fall short of the relaxation bound, but against an exact dynamic program on all $30$ SD-v1.5 subjects it retains $99.997\%$ of the DP-optimal energy at a $5$\,MiB payload and $99.999\%$ at $13$\,MiB. The guarantee concerns energy, not DINO; Sections~\ref{sec:adaptive} and~\ref{sec:law} measure what the extra energy yields in fidelity.

\paragraph{Cost of compression.}
One pass of per-layer SVDs takes, excluding weight loading, from $42$\,s on SD-v1.5 to $592$\,s on FLUX.1-dev on a single A100, orders of magnitude less than the fine-tuning that produced the checkpoint.

\section{Experimental Setup}
\label{sec:setup}

\paragraph{Subjects and backbones.}
The controlled study uses the $30$ DreamBooth subjects~\citep{ruiz2023dreambooth} on SD-v1.5, each fully fine-tuned; every method is evaluated on the same subjects, so all comparisons are paired. We add $84$ CustomConcept101 subjects~\citep{kumari2023customdiffusion}, which include the humans, vehicles, and scenes the DreamBooth set lacks (Section~\ref{sec:cc101}), and train SD3.5-medium~\citep{esser2024sd3,stabilityai2024sd35}, a $2.2$B MMDiT-X, on $12$ subjects (Section~\ref{sec:sd3}). SDXL~\citep{podell2023sdxl} and FLUX.1-dev~\citep{flux2024} are personalized with a high-rank LoRA merged into the base, the strongest fine-tuning that fits in $80$\,GB at those sizes; since the merge bounds the delta's rank, those results are reported separately (Section~\ref{sec:merged}).

\paragraph{Baselines.}
\textbf{\textcolor{gpb}{Group B}} receives only the finished checkpoint, which is our setting: whole-model quantization of the personalized UNet to 8 and 4 bits (LLM.int8~\citep{dettmers2022llmint8} and NF4~\citep{dettmers2023qlora} via \texttt{bitsandbytes}); int8 and int4 quantization of the dense delta $\dW$ with per-row scales; a 1-bit sign-and-scale delta, the closed-form initialization of BitDelta~\citep{liu2024bitdelta} without its distillation stage; magnitude pruning of $\dW$ at matched sparsity; a random-rank ablation; and \dsvd{}, all scored under one protocol on all $30$ subjects. \textbf{\textcolor{gpa}{Group A}} trains from scratch with the subject images: full fine-tuning, LoRA at ranks $4$, $8$, and $16$, and an SVDiff~\citep{han2023svdiff} adapter with tuned learning rate and step count; Section~\ref{sec:exp} adds a LoRA learning-rate sweep on six subjects.

\paragraph{Metrics and statistics.}
Following the DreamBooth protocol we report \textbf{CLIP-I}~\citep{radford2021clip} and \textbf{DINO}~\citep{caron2021dino} similarity to the subject's reference images, \textbf{CLIP-T} for prompt following, and FID~\citep{heusel2017fid} and KID~\citep{binkowski2018kid}; DINO separates methods most clearly and is our headline metric. Reconstruction fidelity, written DINO$\to$DB (likewise FID$\to$DB, KID$\to$DB), is DINO between a compressed model's image and the uncompressed model's image for the same prompt and seed. Aggregates carry bootstrap $95\%$ intervals; differences use paired Wilcoxon tests~\citep{wilcoxon1945} and equivalence TOST at a margin of $\pm0.01$ DINO; given the number of tests we treat $p<10^{-3}$ as robust and flag near-$0.05$ results that would not survive a Holm correction.

\paragraph{Storage accounting.}
Sizes are measured on disk, in MiB ($2^{20}$ bytes) for SD-v1.5 and in decimal MB ($10^6$ bytes) for SD3.5-medium, SDXL, and FLUX.1. The uncompressed SD-v1.5 pipeline is $4{,}994$\,MiB, of which the fp32 UNet is $3{,}279$\,MiB and an fp16 UNet $1{,}639$\,MiB; ratios against $4{,}994$\,MiB describe replacing whole checkpoints, and Section~\ref{sec:deploy} gives the shared-base accounting against the UNet payload.

\paragraph{Implementation details.}
Images are generated with \texttt{diffusers}~\citep{vonplaten2022diffusers} $0.31.0$ at $50$ steps, $100$ per subject under fixed seeds and the standard prompt template, plus a $25$-prompt editability suite for CLIP-T. DINO uses \texttt{facebook/dino-vits16}. Latency and VRAM are means over $5$ runs after $2$ warm-ups; all timings use a single A100 ($80$\,GB).

\section{Results on the 30-Subject Benchmark}
\label{sec:exp}

\subsection{Fidelity along the energy ladder}
At $\tau{=}0.5$ the adapter averages $708$\,MiB against the $4{,}994$\,MiB checkpoint ($7\times$) and is equivalent to the uncompressed model within the $\pm0.01$ DINO margin: DINO $0.706$ against $0.707$, TOST $p{=}9{\times}10^{-10}$. Fidelity stays level down to $\tau{=}0.2$ ($151$\,MiB, $33\times$), which keeps $98\%$ of the uncompressed DINO and which we call the knee. Below it identity remains recognizable at $\tau{=}0.1$ ($55$\,MiB, DINO $0.662$) and degrades at $\tau{=}0.01$ ($5.5$\,MiB, $0.577$), where a mean per-layer rank of one still preserves the class and coarse appearance (Figure~\ref{fig:qualitative}). CLIP-T rises mildly as $\tau$ falls (Figure~\ref{fig:pareto}, right), because a truncated delta overrides the prompt less; \dsvd{} moves along the identity--editability trade-off smoothly, whereas pruning and the 1-bit delta fall below it.

\subsection{Post-hoc baselines above \texorpdfstring{$400$\,MiB}{400 MiB}}
Quantizing the whole personalized UNet is faithful (LLM.int8: DINO $0.707$; NF4: $0.702$) but barely compresses ($3{,}354$ and $2{,}586$\,MiB) and slows inference (Section~\ref{sec:deploy}); \dsvd{} at $\tau{=}0.5$ reaches the same fidelity in $3.7$--$4.7\times$ less space. Quantizing the delta is far more effective: dense $\dW$ at int8 ($821$\,MiB) and int4 ($411$\,MiB) is lossless, with a paired DINO deficit below $0.001$, TOST $p{<}10^{-23}$, and DINO$\to$DB $0.833$. Above about $400$\,MiB scalar delta quantization is therefore the method of choice, and \dsvd{} at $\tau{=}0.5$ offers no size advantage over int4.

\subsection{Post-hoc baselines below \texorpdfstring{$400$\,MiB}{400 MiB}}
The 1-bit sign-and-scale delta does not hold: $103$\,MiB at DINO $0.555$, a deficit of $0.153$, where \dsvd{} scores $0.695$ ($151$\,MiB) and $0.662$ ($55$\,MiB) and continues down to $5.5$\,MiB. Magnitude pruning at $99\%$ sparsity ($50$\,MiB) falls to $0.361$ where \dsvd{} scores $0.66$, and keeping $6$ random singular components instead of the top $6$ costs $0.33$ DINO at identical storage (Table~\ref{tab:adaptive}): the subject lies in the leading directions of each layer's spectrum. The families combine: int4 factors shrink the $5.5$\,MiB adapter a further $2.2\times$ to $2.5$\,MiB with no measurable change (Table~\ref{tab:main}). Below $400$\,MiB, \dsvd{} is thus the only post-hoc method we measured that preserves identity.

\subsection{Trained adapters, when the data exist}
With the subject images, LoRA is strong at a few MiB: $r{=}4$ reaches DINO $0.668$ at $3.1$\,MiB and $r{=}8$ reaches $0.651$ at $6.1$\,MiB, against $0.577$ for \dsvd{} at $5.5$\,MiB; tuned SVDiff scores $0.606$ at $0.88$\,MiB, $0.027$ above our $5.5$\,MiB point ($p{=}0.02$, not significant after Holm correction). At the $10$-MiB scale tuning decides: the $30$-subject $r{=}16$ row of Table~\ref{tab:main} scores $0.608$ at one learning rate, and a sweep on six subjects (Table~\ref{tab:lora}) raises $r{=}16$ from $0.565$ to $0.627$ and $r{=}8$ to $0.674$, at or above \dsvd{} rank $6$ at $13.1$\,MiB, which scores $0.601$ on the same six ($r{=}16$ and rank $6$ are indistinguishable there); a tuned LoRA thus matches or beats post-hoc truncation at this size. Against SVDiff the ordering reverses as size grows: \dsvd{} leads by $+0.051$ DINO at $55$\,MiB and $+0.078$ at $708$\,MiB. No trained adapter reproduces the checkpoint: the best LoRA reaches DINO$\to$DB $0.707$, whereas \dsvd{} reaches $0.77$ at $13$\,MiB and $0.81$--$0.83$ from the knee upward, which is what matters when the images are gone or an existing checkpoint's behavior must be preserved.

\begin{table}[t]
\centering
\caption{\textbf{30-subject benchmark, SD-v1.5.} CLIP-I/DINO/CLIP-T and FID/KID (pooled over $3{,}000$ images per variant; KID is unbiased and can be negative) score against subject references; DINO$\to$DB scores reconstruction of the uncompressed model's outputs. \textcolor{gpa}{Group A} trains adapters with data access; \textcolor{gpb}{Group B} compresses the finished checkpoint. Sizes in MiB on disk. ``Whole-model'' rows quantize the personalized UNet with \texttt{bitsandbytes}; ``dense $\dW$'' rows quantize the delta with per-row scales (int8/int4) or per-matrix sign-and-scale (1-bit). Best Group-B value per column in \textbf{bold}; knee shaded.}
\label{tab:main}
\footnotesize
\setlength{\tabcolsep}{2.5pt}
\begin{tabular}{llccccccc}
\toprule
& Method & CLIP-I$\uparrow$ & DINO$\uparrow$ & CLIP-T$\uparrow$ & FID$\downarrow$ & KID$\downarrow$ & DINO$\to$DB$\uparrow$ & Size (MiB)$\downarrow$ \\
\midrule
& Full fine-tune (uncompressed) & 0.838 & 0.707 & 0.219 & 126.9 & 4.42 & ---   & 4994 \\
\midrule
\multirow{4}{*}{\rotatebox{90}{\textcolor{gpa}{Group A}}}
& LoRA $r{=}4$            & 0.798 & 0.668 & 0.214 & 118.9 & 3.89 & 0.707 & 3.1  \\
& LoRA $r{=}8$            & 0.778 & 0.651 & 0.210 & 117.4 & 4.93 & 0.695 & 6.1  \\
& LoRA $r{=}16$           & 0.759 & 0.608 & 0.207 & 116.9 & 6.85 & 0.643 & 12.2 \\
& SVDiff (tuned)          & 0.786 & 0.606 & 0.225 & 145.5 & 5.32 & 0.668 & 0.88 \\
\midrule
\multirow{14}{*}{\rotatebox{90}{\textcolor{gpb}{Group B (post-hoc)}}}
& Whole-model int8 (LLM.int8)   & 0.837 & \textbf{0.707} & 0.219 & 127.2 & 4.40 & 0.830 & 3354 \\
& Whole-model 4-bit (NF4)       & 0.835 & 0.702 & 0.220 & 127.0 & \textbf{3.79} & 0.808 & 2586 \\
& Dense $\dW$ int8              & 0.838 & \textbf{0.707} & 0.219 & ---   & ---  & \textbf{0.833} & 821.0 \\
& Dense $\dW$ int4              & 0.838 & \textbf{0.707} & 0.219 & ---   & ---  & \textbf{0.833} & 411.4 \\
& Dense $\dW$ 1-bit (BitDelta init.) & 0.754 & 0.555 & 0.232 & --- & --- & 0.644 & 103.3 \\
& Prune $s{=}0.99$        & 0.659 & 0.361 & 0.236 & 163.6 & 7.89 & 0.424 & 50.0 \\
& Random rank $6$         & 0.654 & 0.347 & 0.235 & 166.1 & 8.46 & 0.409 & 13.1 \\
& \dsvd{} $\tau{=}0.01$   & 0.765 & 0.577 & 0.232 & 143.6 & 5.71 & 0.655 & 5.5 \\
& \;\; + int4 factors     & 0.766 & 0.577 & 0.231 & ---   & ---  & 0.655 & \textbf{2.5} \\
& \dsvd{} rank $6$        & 0.824 & 0.677 & 0.223 & 135.6 & 4.71 & 0.775 & 13.1 \\
& \dsvd{} budget $13$\,MiB & 0.823 & 0.677 & 0.223 & 135.5 & 4.71 & 0.773 & 13.5 \\
& \dsvd{} $\tau{=}0.1$    & 0.819 & 0.662 & 0.224 & 136.9 & 5.15 & 0.765 & 55.3 \\
\rowcolor{gray!15}
& \dsvd{} $\tau{=}0.2$ (knee) & 0.831 & 0.695 & 0.220 & 130.4 & 4.50 & 0.809 & 150.9 \\
& \dsvd{} $\tau{=}0.5$    & \textbf{0.839} & 0.706 & 0.219 & \textbf{127.6} & 4.44 & 0.829 & 708.3 \\
\bottomrule
\end{tabular}
\end{table}

\begin{figure}[t]
\centering
\begin{minipage}[t]{0.44\textwidth}
\centering
\includegraphics[width=\linewidth]{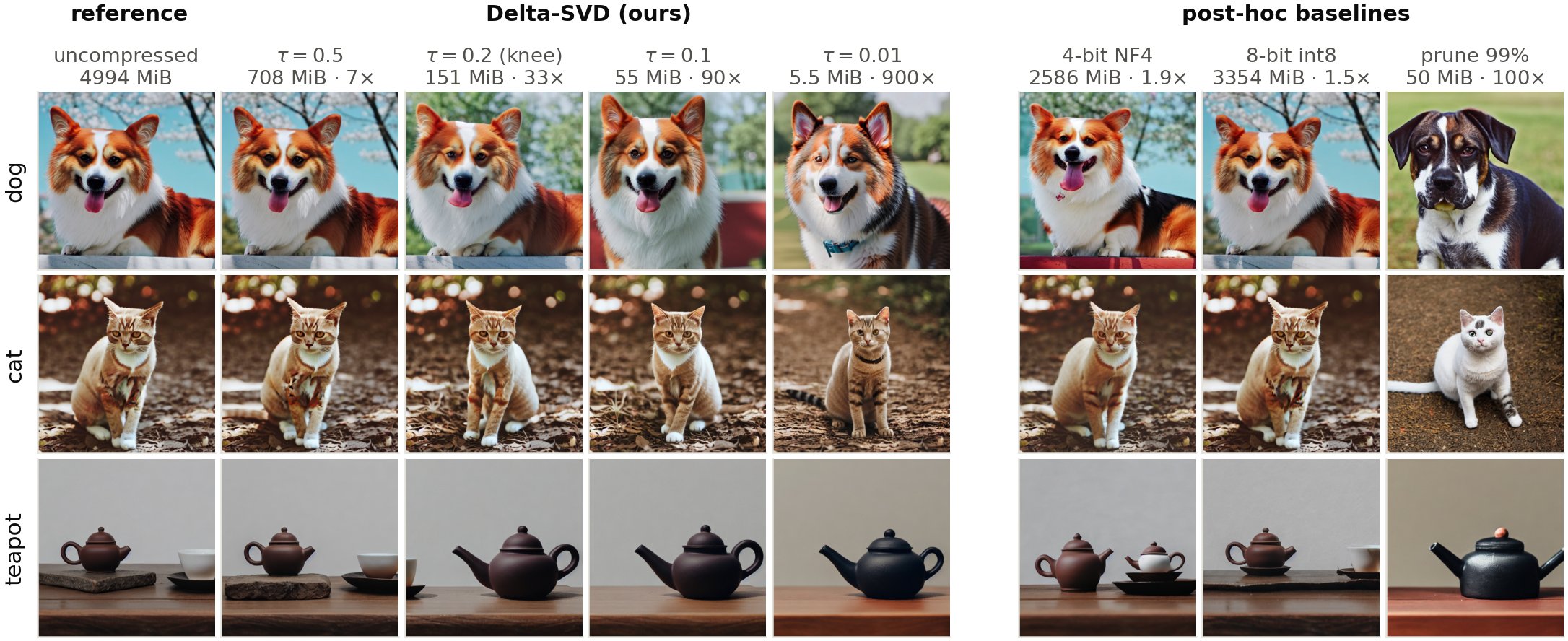}
\caption{\textbf{Matched-seed samples on SD-v1.5} (same subject, prompt, and seed per row). Identity survives to $\tau{=}0.1$ ($90\times$) and drifts only at $\tau{=}0.01$; whole-model NF4 and LLM.int8 are faithful but barely compress, and $99\%$ pruning returns a generic class instance.}
\label{fig:qualitative}
\end{minipage}\hfill
\begin{minipage}[t]{0.48\textwidth}
\centering
\includegraphics[width=\linewidth]{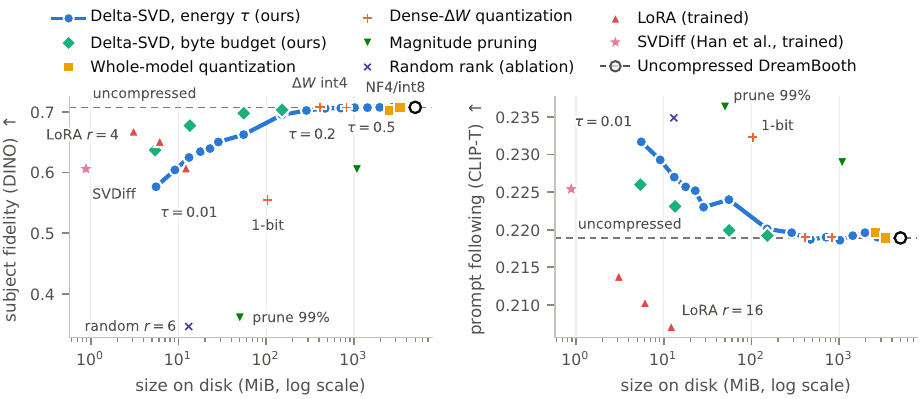}
\caption{\textbf{Storage--quality frontier} (log-scale MiB); left: subject fidelity, right: prompt following. Dense-delta int8/int4 are at the uncompressed level at $411$--$821$\,MiB; below that the \dsvd{} ladder and budget allocator trace the upper envelope over two orders of magnitude, while the 1-bit delta, whole-model quantization, pruning, and random-rank selection fall below it. Trained adapters compete in the few-MiB corner.}
\label{fig:pareto}
\end{minipage}
\end{figure}

\subsection{Allocator versus fixed threshold}
\label{sec:adaptive}
At matched bytes (Table~\ref{tab:adaptive}: paired single-prompt DINO, the $25$-prompt suite, and pooled FID$\to$DB) the allocator gains $+0.056$ DINO at a tight $5$\,MiB budget ($p{=}3.5{\times}10^{-8}$); the margin shrinks to $+0.016$ at $55$\,MiB and $+0.006$ at $151$\,MiB as saturated spectra leave nothing to reallocate. It ties uniform rank $6$, and both are far above random selection. Most of the value thus comes from keeping the top of each spectrum; since the allocator is within $0.003\%$ of the integer energy optimum (Section~\ref{sec:budget}), rank allocation has little further headroom on this backbone.

\begin{table}[t]
\centering
\caption{\textbf{Budget water-filling vs.\ fixed-$\tau$ thresholding at matched storage} (paired over $30$ subjects; positive favors the allocator).}
\label{tab:adaptive}
\footnotesize
\begin{tabular}{lccc}
\toprule
Matched pair & $\Delta$DINO (single-prompt) & $\Delta$DINO (suite) & pooled FID$\to$DB \\
\midrule
budget $5$\,MiB vs.\ $\tau{=}0.01$   & $+0.056$ ($p{=}3.5{\times}10^{-8}$) & $+0.026$ & $19.2 < 22.2$ \\
budget $55$\,MiB vs.\ $\tau{=}0.1$   & $+0.016$ ($p{=}4.7{\times}10^{-7}$) & $+0.029$ & $11.1 < 14.7$ \\
budget $151$\,MiB vs.\ $\tau{=}0.2$  & $+0.006$ ($p{=}1.7{\times}10^{-4}$) & $+0.007$ & $8.6 < 10.5$ \\
budget $13$\,MiB vs.\ rank $6$       & $\pm 0$ ($p{=}0.43$) & tie & $15.7 \approx 15.5$ \\
budget $13$\,MiB vs.\ random $6$     & $+0.32$ ($p{=}1.9{\times}10^{-9}$) & $+0.18$ & $15.7 \ll 42.0$ \\
\bottomrule
\end{tabular}
\end{table}

\begin{table}[t]
\centering
\caption{\textbf{LoRA learning-rate sweep} on six DreamBooth subjects (backpack, backpack\_dog, bear\_plushie, berry\_bowl, can, candle), DINO vs.\ references under the Table~\ref{tab:main} protocol; the $30$-subject $r{=}16$ row of Table~\ref{tab:main} scores $0.565$ on these six.}
\label{tab:lora}
\footnotesize
\begin{tabular}{llcc}
\toprule
Method & lr & MiB & DINO$\uparrow$ \\
\midrule
LoRA $r{=}16$ & $2.5\mathrm{e}{-5}$ & 12.20 & 0.619 \\
LoRA $r{=}16$ & $1\mathrm{e}{-4}$   & 12.20 & 0.627 \\
LoRA $r{=}16$ & $4\mathrm{e}{-4}$   & 12.20 & 0.522 \\
LoRA $r{=}8$  & $1\mathrm{e}{-4}$   & \phantom{0}6.12 & 0.674 \\
\dsvd{} $r{=}6$ (no data) & --- & 13.07 & 0.601 \\
\bottomrule
\end{tabular}
\end{table}

\subsection{Energy predicts fidelity}
\label{sec:law}
Across the $\tau$ ladder, reconstruction fidelity is nearly a function of the preserved $\sigma^2$-energy: variant means follow it at $\rho{=}1.00$ and $R^2{=}0.87$. Inverting the fit gives the energy, and hence the size, required for a target fidelity: a target DINO$\to$DB of $0.80$ inverts to energy $0.77$, which the spectra place at $\tau{\approx}0.3$ (about $290$\,MiB), and the measured $\tau{=}0.3$ adapter scores $0.823$. Adding the budget and random variants lowers $R^2$ to $0.42$: equal-energy allocations with different layer placement are not equally good, the slack the allocator exploits.

\section{Generality}
\label{sec:general}

We extend Section~\ref{sec:exp} to more diverse subjects and a second fully fine-tuned backbone, then report separately on checkpoints whose delta was rank-bounded by training.

\subsection{Subject diversity: 84 CustomConcept101 subjects}
\label{sec:cc101}
\begin{figure}[t]
\centering
\begin{minipage}[b]{0.31\textwidth}
\centering
\includegraphics[width=\linewidth]{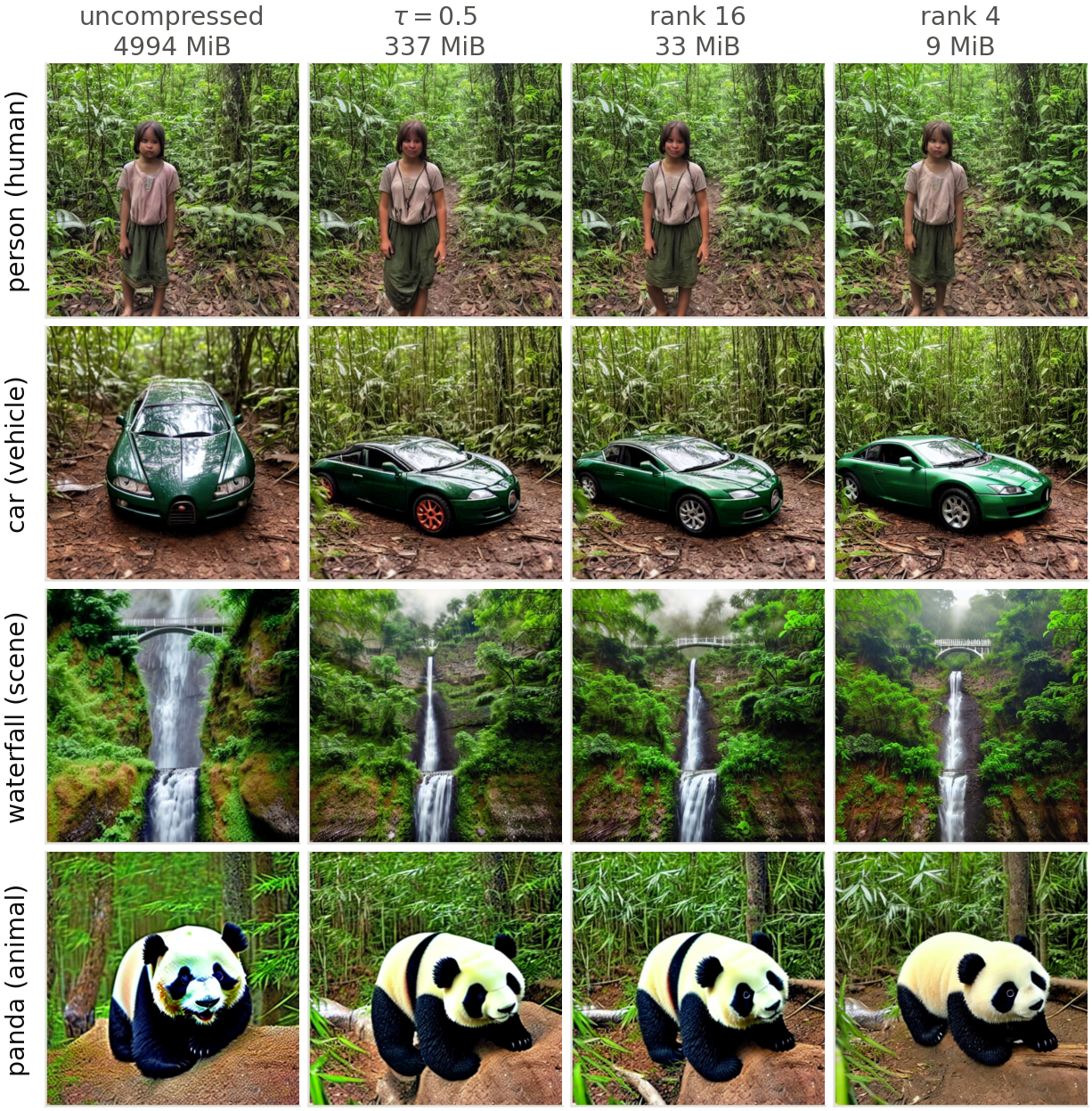}
\end{minipage}\hfill
\begin{minipage}[b]{0.40\textwidth}
\centering
\includegraphics[width=\linewidth]{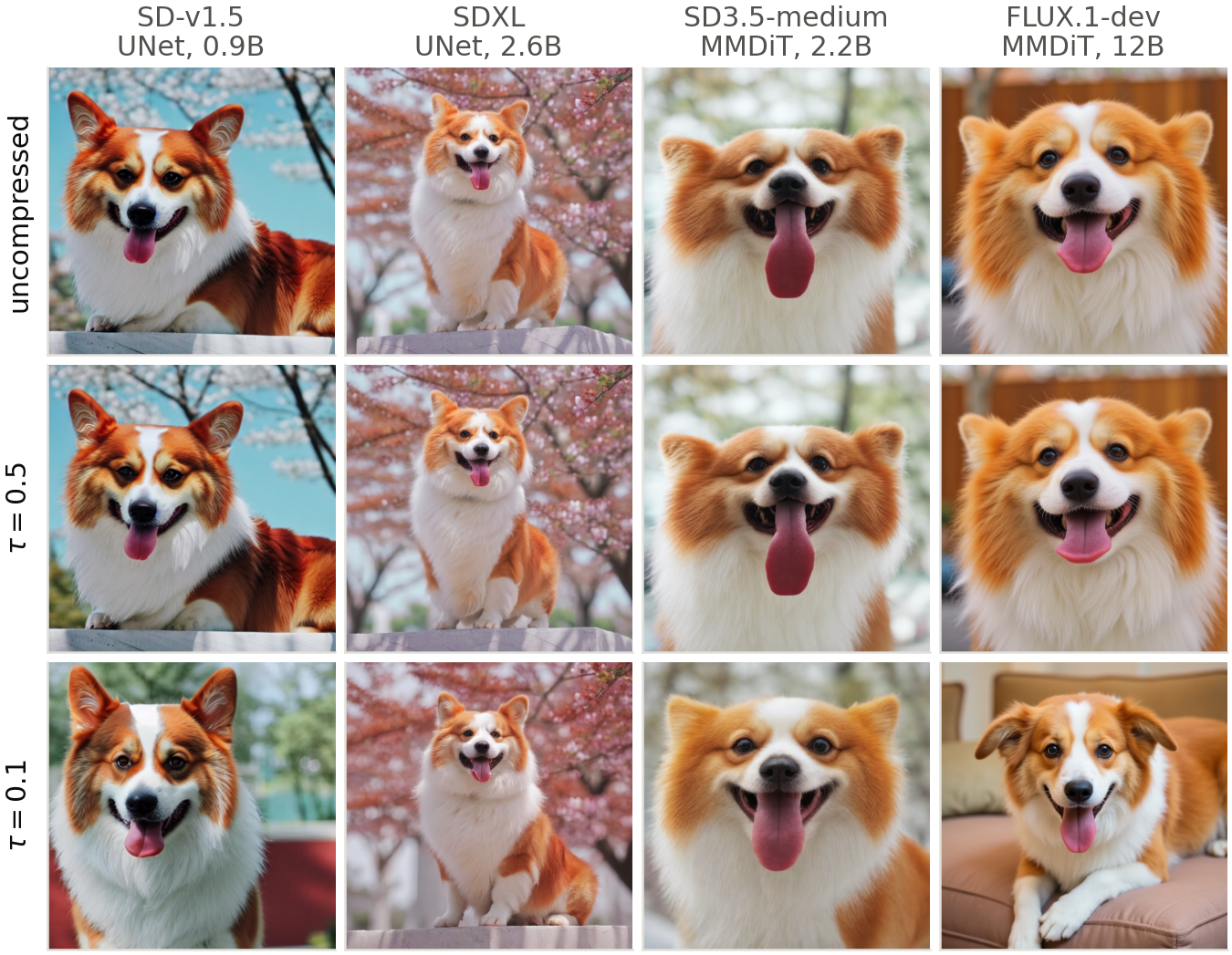}
\end{minipage}
\caption{\textbf{Left: CustomConcept101 reconstruction} at matched prompt and seed (``in the jungle''), one subject per category; $337$\,MiB to $9$\,MiB preserves subject and composition across humans, vehicles, scenes, and animals. \textbf{Right: reconstruction across backbones} (columns) at three settings (rows); SD-v1.5 and SD3.5-medium fully fine-tuned, SDXL and FLUX.1 merged LoRAs. At $\tau{=}0.5$ each backbone's personalized output is reproduced almost exactly; at $\tau{=}0.1$ identity persists while detail and composition drift, most visibly on FLUX.1.}
\label{fig:cc101}\label{fig:crossqual}
\end{figure}

On $84$ CustomConcept101 subjects~\citep{kumari2023customdiffusion} (SD-v1.5), Table~\ref{tab:cc101} gives three readings. Against reference images on the $25$-prompt suite all variants score DINO $0.40$ to $0.43$; this level belongs to the backbone, not to compression, since on the $12$ subjects whose uncompressed checkpoints we kept the uncompressed model scores $0.532$ against $0.531$, $0.527$, and $0.517$ for the $337$, $33$, and $9$\,MiB adapters, a loss of at most $0.015$ DINO even on people and scenes. Paired reconstruction at the same prompt and seed is the sharper reading: the $9$\,MiB rank-$4$ adapter reproduces the uncompressed outputs at DINO $0.770$, near the $0.787$ of the $337$\,MiB adapter, with per-category means from $0.756$ (people) to $0.804$ (vehicles); Figure~\ref{fig:cc101} (left) shows samples. The low-rank premise and the identity--editability trade-off are thus properties of the fine-tuning, not of the subject category.

\begin{table}[t]
\centering
\caption{\textbf{CustomConcept101 subject-diversity study} (SD-v1.5, $25$-prompt suite $\times$ $4$ seeds). Top: $84$ subjects against reference images (the streaming pipeline deletes each checkpoint after compression, so no uncompressed column exists at $n{=}84$). Middle: the $12$ subjects (3 human, 4 vehicle, 4 scene, 1 animal) whose uncompressed checkpoints were kept. Bottom: same-seed paired reconstruction against those outputs, with per-category paired DINO at $\tau{=}0.5$ (the single-subject animal cell is illustrative only).}
\label{tab:cc101}
\footnotesize
\setlength{\tabcolsep}{5pt}
\begin{tabular}{lcccc}
\toprule
 & Uncompressed & \dsvd{} rank $4$ & \dsvd{} rank $16$ & \dsvd{} $\tau{=}0.5$ \\
\midrule
Size (MiB) & 4994 & 9.0 & 33.2 & 336.8 \\
\midrule
\multicolumn{5}{l}{\emph{vs.\ reference images, all subjects ($n{=}84$)}} \\
DINO & --- & 0.401 & 0.417 & 0.429 \\
CLIP-I & --- & 0.641 & 0.653 & 0.664 \\
CLIP-T & --- & 0.263 & 0.259 & 0.255 \\
\midrule
\multicolumn{5}{l}{\emph{vs.\ reference images, subset with uncompressed checkpoint kept ($n{=}12$)}} \\
DINO & 0.532 & 0.517 & 0.527 & 0.531 \\
CLIP-I & 0.698 & 0.667 & 0.679 & 0.686 \\
\midrule
\multicolumn{5}{l}{\emph{paired reconstruction vs.\ uncompressed outputs ($n{=}12$, same seed)}} \\
Paired DINO & --- & 0.770 & 0.787 & 0.787 \\
\midrule
\multicolumn{2}{l}{Per category, $\tau{=}0.5$:} & \multicolumn{3}{l}{human $0.756$ \;·\; vehicle $0.804$ \;·\; scene $0.797$ \;·\; animal $0.779$} \\
\bottomrule
\end{tabular}
\end{table}

\subsection{A second fully fine-tuned architecture: SD3.5-medium}
\label{sec:sd3}
On the fully fine-tuned SD3.5-medium MMDiT ($12$ subjects, same recipe, eight matched-seed images per subject and variant) truncation behaves as on the UNet (Table~\ref{tab:sd3}). At $\tau{=}0.5$ the adapter stores $531$\,MB, matches the uncompressed model on every reference metric (DINO $0.742$ against $0.743$), and reproduces its outputs at DINO$\to$DB $0.890$; $\tau{=}0.9$ adds $0.002$ at $5.6\times$ the size, so $\tau{=}0.5$ is again the equivalence point. At $\tau{=}0.1$ ($31$\,MB) the DINO deficit is $0.009$, against $0.045$ for the UNet, and identity degrades only at $\tau{=}0.01$ ($6.9$\,MB, deficit $0.082$); Figure~\ref{fig:crossqual} (right) shows samples.

\begin{table}[t]
\centering
\caption{\textbf{SD3.5-medium (MMDiT, full fine-tuning, $12$ subjects, $8$ matched-seed images per subject and variant).} Left: scores against reference images; right: reconstruction against the uncompressed model's outputs. Sizes in decimal MB; as on the UNet, $\tau{=}0.5$ (shaded) is equivalent to the uncompressed model.}
\label{tab:sd3}
\footnotesize
\setlength{\tabcolsep}{5pt}
\begin{tabular}{lcccccc}
\toprule
Variant & Size (MB) & DINO$\uparrow$ & CLIP-I$\uparrow$ & CLIP-T$\uparrow$ & DINO$\to$DB$\uparrow$ & CLIP-I$\to$DB$\uparrow$ \\
\midrule
Uncompressed & --- & 0.743 & 0.876 & 0.214 & --- & --- \\
\dsvd{} $\tau{=}0.9$  & 2952.8 & 0.743 & 0.875 & 0.213 & 0.892 & 0.932 \\
\rowcolor{gray!15}
\dsvd{} $\tau{=}0.5$  & 531.4 & 0.742 & 0.874 & 0.213 & 0.890 & 0.931 \\
\dsvd{} $\tau{=}0.1$  & 31.2 & 0.734 & 0.854 & 0.215 & 0.864 & 0.908 \\
\dsvd{} $\tau{=}0.01$ & 6.9 & 0.661 & 0.800 & 0.228 & 0.759 & 0.844 \\
\bottomrule
\end{tabular}
\end{table}

\subsection{Merged-LoRA checkpoints: SDXL and FLUX.1}
\label{sec:merged}
SDXL and FLUX.1-dev ($12$B) do not fit full fine-tuning in $80$\,GB, so their checkpoints come from a high-rank LoRA ($r\le64$ for SDXL, $r{=}256$ for FLUX.1) merged into the base, which bounds the rank of $\dW$ by construction. The measured spectra lie far below that bound (FLUX.1 stable rank $1.48$), but a low-rank delta is what the recipe produces, so these figures say nothing about architecture or scale. Table~\ref{tab:merged} shows that \dsvd{} compresses a merged adapter further with the ordering of Section~\ref{sec:exp}: fidelity to each backbone's uncompressed outputs rises with $\tau$ to DINO$\to$DB $0.86$ at $\tau{=}0.5$ on both, at sizes that are small because the delta was rank-bounded first. On FLUX.1 the near-rank-one spectrum makes $\tau{=}0.01$ and $\tau{=}0.1$ almost the same adapter (Figure~\ref{fig:crossqual}, right).

\begin{table}[t]
\centering
\caption{\textbf{Further compression of merged-LoRA checkpoints} (DINO$\to$DB / CLIP-I$\to$DB against each backbone's uncompressed personalized outputs; $30$ subjects for SDXL, $6$ for FLUX.1, indicative).}
\label{tab:merged}
\footnotesize
\setlength{\tabcolsep}{4pt}
\begin{tabular}{llccc}
\toprule
Backbone & Variant & Size (MB) & DINO$\to$DB$\uparrow$ & CLIP-I$\to$DB$\uparrow$ \\
\midrule
\multirow{3}{*}{SDXL UNet, merged LoRA $r{\le}64$}
 & $\tau{=}0.1$ & 3.8 & 0.798 & 0.893 \\
 & $\tau{=}0.3$ & 9.7 & 0.834 & 0.910 \\
 & $\tau{=}0.5$ & 22.4 & 0.860 & 0.923 \\
\midrule
\multirow{3}{*}{FLUX.1-dev MMDiT, merged LoRA $r{=}256$}
 & $\tau{=}0.01$ & 8.5 & 0.779 & 0.853 \\
 & $\tau{=}0.1$  & 9.0 & 0.780 & 0.852 \\
 & $\tau{=}0.5$  & 440.9 & 0.856 & 0.931 \\
\bottomrule
\end{tabular}
\end{table}

\section{Deployment Analysis}
\label{sec:deploy}

\paragraph{Inference cost.}
\dsvd{} reconstructs dense weights at load time \eqref{eq:merge}, so per-image latency and peak VRAM equal those of the uncompressed model at every $\tau$: $2.28$--$2.34$\,s per image and about $2.7$\,GB. Quantized serving dequantizes in every forward pass: $3.39$\,s per image with NF4 ($1.5\times$) and $7.86$\,s with LLM.int8 ($3.4\times$) in the same \texttt{bitsandbytes} setup. The dense-delta int8/int4 rows of Table~\ref{tab:main} are likewise merged at load.

\paragraph{Load time.}
Table~\ref{tab:loading} separates two regimes. A \emph{base-resident} worker keeps the base weights and switches subject by reading and applying one artifact: with the page cache evicted, a $\tau{=}0.2$ adapter loads in $0.40$\,s against $8.4$\,s for the full fp32 UNet, a $21\times$ speedup, and $\tau{=}0.01$ reaches $88\times$. Applying the delta takes about $30$\,ms up to $\tau{=}0.5$, so the saving is almost entirely artifact I/O, and with a warm cache every setting loads in under a second. A \emph{fresh worker} must also build the pipeline and load the base, which dominates: evicted startup is $6.7$\,s for the full pipeline against $5.8$\,s at $\tau{=}0.2$ and $7.6$\,s at $\tau{=}0.5$. \dsvd{} therefore speeds up subject switching on base-resident workers, the multi-tenant case, and is neutral for cold starts.

\begin{table}[t]
\centering
\caption{\textbf{Loading cost in two regimes} (SD-v1.5, subject \emph{cat}, A100). Base-resident swap: artifact read plus delta reconstruction and application (``apply'', included in the evicted time) with the base in memory; evicted values average trials 2--3 after page-cache eviction, warm values three trials. Fresh-worker startup: three-trial median; at $\tau{=}0.9$ it exceeds the full pipeline because the worker reads the base and a $2.7$\,GB artifact, and other differences are within the one-to-three-second spread of pipeline construction.}
\label{tab:loading}
\footnotesize
\setlength{\tabcolsep}{3pt}
\begin{tabular}{lccccccc}
\toprule
& \multicolumn{4}{c}{Base-resident swap} & & \multicolumn{2}{c}{Fresh-worker startup (median)} \\
\cmidrule{2-5}\cmidrule{7-8}
Variant & evicted (s) & apply (s) & speedup & warm (s) & & evicted (s) & warm (s) \\
\midrule
Full UNet / pipeline & 8.435 & 0.020 & $1.0\times$  & 0.736 & & 6.729 & 0.985 \\
\dsvd{} $\tau{=}0.01$ & 0.096 & 0.031 & $87.8\times$ & 0.090 & & 7.454 & 1.998 \\
\dsvd{} $\tau{=}0.1$  & 0.194 & 0.032 & $43.5\times$ & 0.104 & & ---   & ---   \\
\dsvd{} $\tau{=}0.2$  & 0.395 & 0.031 & $21.3\times$ & 0.124 & & 5.822 & 1.362 \\
\dsvd{} $\tau{=}0.5$  & 1.747 & 0.038 & $4.8\times$  & 0.254 & & 7.623 & 1.297 \\
\dsvd{} $\tau{=}0.9$  & 6.798 & 0.092 & $1.2\times$  & 0.755 & & 14.865 & 1.814 \\
\bottomrule
\end{tabular}
\end{table}

\paragraph{Catalog storage.}
A catalog of $N$ dense checkpoints costs $N\,S_{\text{full}}$; with \dsvd{} it costs $S_0 + N\,S_\Delta$ (VAE and text encoders counted once in both). Against the UNet payload, the $151$\,MiB $\tau{=}0.2$ adapter reduces per-subject storage by $10.9\times$ (fp16 UNet, $1{,}639$\,MiB) or $21.7\times$ (fp32, $3{,}279$\,MiB); the $33\times$ quoted earlier replaces whole $4{,}994$\,MiB checkpoints. For $N{=}1{,}000$ SD-v1.5 models, full checkpoints occupy $4.8$\,TiB, \dsvd{} at $\tau{=}0.5$ $0.68$\,TiB, and the $\tau{=}0.2$ knee $0.14$\,TiB at $98\%$ fidelity; at $5.5$\,MiB per subject the catalog is $5.4$\,GiB and fits in one node's memory. Conversion costs $42$\,s per subject, once, about $12$ GPU-hours for the catalog.

\section{Conclusion}
\label{sec:conclusion}

Full-model personalization leaves a weight delta that is spectrally concentrated: over $126$ fine-tuned subjects, $90\%$ of its energy lies in $19\%$ of the spectrum on the SD-v1.5 UNet and in under $8\%$ on a fully fine-tuned SD3.5-medium MMDiT. \dsvd{} truncates the delta's SVD under an energy target or a byte budget, stores the factors, and merges them at load time. It matches uncompressed fidelity at $7\times$ compression and holds $98\%$ at $33\times$ with no inference cost, for finished checkpoints without their data. Scalar quantization of the dense delta is the better post-hoc choice above about $400$\,MiB and \dsvd{} provides the identity-preserving operating points below it; on a base-resident worker it also makes subject switching $21\times$ faster at the knee.

Several limitations qualify these claims. The truncation is classical, the contribution is the study around it, and a tuned LoRA with data access matches or beats post-hoc truncation at the $10$-MiB scale. \dsvd{} cannot repair a poorly personalized checkpoint, only store it smaller. Our metrics are automatic; DINO tracks the visual evidence (Figures~\ref{fig:qualitative} and~\ref{fig:cc101}), but the regime at and below $\tau{=}0.1$ needs a human study, and the $\pm0.01$ equivalence margin is chosen rather than perceptual. The transformer evidence is a $12$-subject panel on one architecture; the SDXL and FLUX.1 results describe further compression of merged adapters, not full fine-tuning at those scales; and a fresh worker gains no startup time, because pipeline construction dominates. Stacking \dsvd{} on a quantized or distilled base, porting BitDelta's distillation to diffusion deltas, and jointly compressing adapter \emph{libraries} across subjects are natural next steps.

{\setlength{\bibsep}{0pt plus 0.3ex}\footnotesize
\bibliography{acml26}}

@inproceedings{rombach2022ldm,
  title     = {High-Resolution Image Synthesis with Latent Diffusion Models},
  author    = {Rombach, Robin and Blattmann, Andreas and Lorenz, Dominik and Esser, Patrick and Ommer, Bj{\"o}rn},
  booktitle = {Proceedings of the IEEE/CVF Conference on Computer Vision and Pattern Recognition (CVPR)},
  pages     = {10684--10695},
  year      = {2022}
}

@inproceedings{ho2020ddpm,
  title     = {Denoising Diffusion Probabilistic Models},
  author    = {Ho, Jonathan and Jain, Ajay and Abbeel, Pieter},
  booktitle = {Advances in Neural Information Processing Systems (NeurIPS)},
  volume    = {33},
  pages     = {6840--6851},
  year      = {2020}
}

@inproceedings{ruiz2023dreambooth,
  title     = {DreamBooth: Fine Tuning Text-to-Image Diffusion Models for Subject-Driven Generation},
  author    = {Ruiz, Nataniel and Li, Yuanzhen and Jampani, Varun and Pritch, Yael and Rubinstein, Michael and Aberman, Kfir},
  booktitle = {Proceedings of the IEEE/CVF Conference on Computer Vision and Pattern Recognition (CVPR)},
  pages     = {22500--22510},
  year      = {2023}
}

@inproceedings{gal2023textualinversion,
  title     = {An Image is Worth One Word: Personalizing Text-to-Image Generation using Textual Inversion},
  author    = {Gal, Rinon and Alaluf, Yuval and Atzmon, Yuval and Patashnik, Or and Bermano, Amit H. and Chechik, Gal and Cohen-Or, Daniel},
  booktitle = {International Conference on Learning Representations (ICLR)},
  year      = {2023}
}

@inproceedings{kumari2023customdiffusion,
  title     = {Multi-Concept Customization of Text-to-Image Diffusion},
  author    = {Kumari, Nupur and Zhang, Bingliang and Zhang, Richard and Shechtman, Eli and Zhu, Jun-Yan},
  booktitle = {Proceedings of the IEEE/CVF Conference on Computer Vision and Pattern Recognition (CVPR)},
  pages     = {1931--1941},
  year      = {2023}
}

@inproceedings{podell2023sdxl,
  title     = {SDXL: Improving Latent Diffusion Models for High-Resolution Image Synthesis},
  author    = {Podell, Dustin and English, Zion and Lacey, Kyle and Blattmann, Andreas and Dockhorn, Tim and M{\"u}ller, Jonas and Penna, Joe and Rombach, Robin},
  booktitle = {International Conference on Learning Representations (ICLR)},
  year      = {2024}
}

@inproceedings{esser2024sd3,
  title     = {Scaling Rectified Flow Transformers for High-Resolution Image Synthesis},
  author    = {Esser, Patrick and Kulal, Sumith and Blattmann, Andreas and Entezari, Rahim and M{\"u}ller, Jonas and Saini, Harry and Levi, Yam and Lorenz, Dominik and Sauer, Axel and Boesel, Frederic and others},
  booktitle = {International Conference on Machine Learning (ICML)},
  year      = {2024}
}

@misc{flux2024,
  title        = {{FLUX.1}},
  author       = {{Black Forest Labs}},
  year         = {2024},
  howpublished = {\url{https://github.com/black-forest-labs/flux}}
}

@misc{stabilityai2024sd35,
  title        = {Stable Diffusion 3.5 Medium},
  author       = {{Stability AI}},
  year         = {2024},
  howpublished = {\url{https://huggingface.co/stabilityai/stable-diffusion-3.5-medium}}
}

@inproceedings{hu2022lora,
  title     = {LoRA: Low-Rank Adaptation of Large Language Models},
  author    = {Hu, Edward J. and Shen, Yelong and Wallis, Phillip and Allen-Zhu, Zeyuan and Li, Yuanzhi and Wang, Shean and Wang, Lu and Chen, Weizhu},
  booktitle = {International Conference on Learning Representations (ICLR)},
  year      = {2022}
}

@inproceedings{han2023svdiff,
  title     = {SVDiff: Compact Parameter Space for Diffusion Fine-Tuning},
  author    = {Han, Ligong and Li, Yinxiao and Zhang, Han and Milanfar, Peyman and Metaxas, Dimitris and Yang, Feng},
  booktitle = {Proceedings of the IEEE/CVF International Conference on Computer Vision (ICCV)},
  pages     = {7323--7334},
  year      = {2023}
}

@inproceedings{zhang2023adalora,
  title     = {{AdaLoRA}: Adaptive Budget Allocation for Parameter-Efficient Fine-Tuning},
  author    = {Zhang, Qingru and Chen, Minshuo and Bukharin, Alexander and Karampatziakis, Nikos and He, Pengcheng and Cheng, Yu and Chen, Weizhu and Zhao, Tuo},
  booktitle = {International Conference on Learning Representations (ICLR)},
  year      = {2023}
}

@article{gu2024sara,
  title   = {SARA: Singular-Value Based Adaptive Low-Rank Adaption},
  author  = {Gu, Jihao and Chen, Shuai and Wang, Zelin and Zhang, Yibo and Gong, Ping},
  journal = {arXiv preprint arXiv:2408.03290},
  year    = {2024}
}

@inproceedings{meng2024pissa,
  title     = {PiSSA: Principal Singular Values and Singular Vectors Adaptation of Large Language Models},
  author    = {Meng, Fanxu and Wang, Zhaohui and Zhang, Muhan},
  booktitle = {Advances in Neural Information Processing Systems (NeurIPS)},
  year      = {2024}
}

@article{feng2024trilora,
  title   = {TriLoRA: Integrating SVD for Advanced Style Personalization in Text-to-Image Generation},
  author  = {Feng, Chengcheng and He, Mu and Tian, Qiuyu and Yin, Haojie and Zhao, Xiaofang and Tang, Hongwei and Wei, Xingqiang},
  journal = {arXiv preprint arXiv:2405.11236},
  year    = {2024}
}

@inproceedings{yeh2024lycoris,
  title     = {Navigating Text-To-Image Customization: From {LyCORIS} Fine-Tuning to Model Evaluation},
  author    = {Yeh, Shih-Ying and Hsieh, Yu-Guan and Gao, Zhidong and Yang, Bernard B. W. and Oh, Giyeong and Gong, Yanmin},
  booktitle = {International Conference on Learning Representations (ICLR)},
  year      = {2024}
}

@inproceedings{vasani2025phlora,
  title     = {PHLoRA: Data-Free Post-Hoc Low-Rank Adapter Extraction from Full-Rank Checkpoint},
  author    = {Vasani, Bhoomit and FitzGerald, Jack and Fang, Anjie and Vaish, Sushmit},
  booktitle = {Findings of the Association for Computational Linguistics: IJCNLP-AACL 2025},
  pages     = {1992--1999},
  year      = {2025}
}

@inproceedings{li2023qdiffusion,
  title     = {Q-Diffusion: Quantizing Diffusion Models},
  author    = {Li, Xiuyu and Liu, Yijiang and Lian, Long and Yang, Huanrui and Dong, Zhen and Kang, Daniel and Zhang, Shanghang and Keutzer, Kurt},
  booktitle = {Proceedings of the IEEE/CVF International Conference on Computer Vision (ICCV)},
  pages     = {17535--17545},
  year      = {2023}
}

@inproceedings{he2023ptqd,
  title     = {PTQD: Accurate Post-Training Quantization for Diffusion Models},
  author    = {He, Yefei and Liu, Luping and Liu, Jing and Wu, Weijia and Zhou, Hong and Zhuang, Bohan},
  booktitle = {Advances in Neural Information Processing Systems (NeurIPS)},
  year      = {2023}
}

@inproceedings{li2024svdquant,
  title     = {SVDQuant: Absorbing Outliers by Low-Rank Components for 4-Bit Diffusion Models},
  author    = {Li, Muyang and Lin, Yujun and Zhang, Zhekai and Cai, Tianle and Li, Xiuyu and Guo, Junxian and Xie, Enze and Meng, Chenlin and Zhu, Jun-Yan and Han, Song},
  booktitle = {International Conference on Learning Representations (ICLR)},
  year      = {2025}
}

@inproceedings{kim2024bksdm,
  title     = {BK-SDM: A Lightweight, Fast, and Cheap Version of Stable Diffusion},
  author    = {Kim, Bo-Kyeong and Song, Hyoung-Kyu and Castells, Thibault and Choi, Shinkook},
  booktitle = {European Conference on Computer Vision (ECCV)},
  year      = {2024}
}

@inproceedings{fang2023diffpruning,
  title     = {Structural Pruning for Diffusion Models},
  author    = {Fang, Gongfan and Ma, Xinyin and Wang, Xinchao},
  booktitle = {Advances in Neural Information Processing Systems (NeurIPS)},
  year      = {2023}
}

@inproceedings{dettmers2023qlora,
  title     = {QLoRA: Efficient Finetuning of Quantized LLMs},
  author    = {Dettmers, Tim and Pagnoni, Artidoro and Holtzman, Ari and Zettlemoyer, Luke},
  booktitle = {Advances in Neural Information Processing Systems (NeurIPS)},
  year      = {2023}
}

@article{eckart1936approximation,
  title     = {The Approximation of One Matrix by Another of Lower Rank},
  author    = {Eckart, Carl and Young, Gale},
  journal   = {Psychometrika},
  volume    = {1},
  number    = {3},
  pages     = {211--218},
  year      = {1936}
}

@inproceedings{radford2021clip,
  title     = {Learning Transferable Visual Models From Natural Language Supervision},
  author    = {Radford, Alec and Kim, Jong Wook and Hallacy, Chris and Ramesh, Aditya and Goh, Gabriel and Agarwal, Sandhini and Sastry, Girish and Askell, Amanda and Mishkin, Pamela and Clark, Jack and Krueger, Gretchen and Sutskever, Ilya},
  booktitle = {International Conference on Machine Learning (ICML)},
  pages     = {8748--8763},
  year      = {2021}
}

@inproceedings{caron2021dino,
  title     = {Emerging Properties in Self-Supervised Vision Transformers},
  author    = {Caron, Mathilde and Touvron, Hugo and Misra, Ishan and J{\'e}gou, Herv{\'e} and Mairal, Julien and Bojanowski, Piotr and Joulin, Armand},
  booktitle = {Proceedings of the IEEE/CVF International Conference on Computer Vision (ICCV)},
  pages     = {9650--9660},
  year      = {2021}
}

@inproceedings{heusel2017fid,
  title     = {GANs Trained by a Two Time-Scale Update Rule Converge to a Local Nash Equilibrium},
  author    = {Heusel, Martin and Ramsauer, Hubert and Unterthiner, Thomas and Nessler, Bernhard and Hochreiter, Sepp},
  booktitle = {Advances in Neural Information Processing Systems (NeurIPS)},
  volume    = {30},
  year      = {2017}
}

@inproceedings{binkowski2018kid,
  title     = {Demystifying MMD GANs},
  author    = {Bi{\'n}kowski, Miko{\l}aj and Sutherland, Danica J. and Arbel, Michael and Gretton, Arthur},
  booktitle = {International Conference on Learning Representations (ICLR)},
  year      = {2018}
}

@misc{vonplaten2022diffusers,
  title        = {Diffusers: State-of-the-Art Diffusion Models},
  author       = {von Platen, Patrick and Patil, Suraj and Lozhkov, Anton and Cuenca, Pedro and Lambert, Nathan and Rasul, Kashif and Davaadorj, Mishig and Wolf, Thomas and others},
  year         = {2022},
  howpublished = {\url{https://github.com/huggingface/diffusers}}
}

@article{wilcoxon1945,
  title   = {Individual Comparisons by Ranking Methods},
  author  = {Wilcoxon, Frank},
  journal = {Biometrics Bulletin},
  volume  = {1},
  number  = {6},
  pages   = {80--83},
  year    = {1945}
}

@inproceedings{liu2024bitdelta,
  title     = {BitDelta: Your Fine-Tune May Only Be Worth One Bit},
  author    = {Liu, James and Xiao, Guangxuan and Li, Kai and Lee, Jason D. and Han, Song and Dao, Tri and Cai, Tianle},
  booktitle = {Advances in Neural Information Processing Systems (NeurIPS)},
  year      = {2024}
}

@inproceedings{yao2025deltazip,
  title     = {DeltaZip: Efficient Serving of Multiple Full-Model-Tuned {LLMs}},
  author    = {Yao, Xiaozhe and Hu, Qinghao and Klimovic, Ana},
  booktitle = {Proceedings of the Twentieth European Conference on Computer Systems (EuroSys)},
  year      = {2025}
}

@inproceedings{ping2024deltacome,
  title     = {Delta-CoMe: Training-Free Delta-Compression with Mixed-Precision for Large Language Models},
  author    = {Ping, Bowen and Wang, Shuo and Wang, Hanqing and Han, Xu and Xu, Yuzhuang and Yan, Yukun and Chen, Yun and Chang, Baobao and Liu, Zhiyuan and Sun, Maosong},
  booktitle = {Advances in Neural Information Processing Systems (NeurIPS)},
  year      = {2024}
}

@inproceedings{dettmers2022llmint8,
  title     = {{LLM.int8()}: 8-bit Matrix Multiplication for Transformers at Scale},
  author    = {Dettmers, Tim and Lewis, Mike and Belkada, Younes and Zettlemoyer, Luke},
  booktitle = {Advances in Neural Information Processing Systems (NeurIPS)},
  year      = {2022}
}

\appendix
\section{Optimality of the Byte-Budget Allocator}
\label{app:proof}

We prove Proposition~\ref{prop:waterfill} and a companion corollary; the objective throughout is retained $\sigma^2$-energy, not DINO (Section~\ref{sec:adaptive}).

\paragraph{Formulation.}
With $v_{li}$, $c_l$, and $x_{li}$ as in Section~\ref{sec:budget}, write $E(x) = \sum_{l,i} v_{li}\, x_{li}$ and $C(x) = \sum_{l,i} c_l\, x_{li}$. The relaxation of \eqref{eq:budget} is $\max_x E(x)$ subject to $C(x) \le B$ and $1 \ge x_{l1} \ge x_{l2} \ge \cdots \ge 0$ for every $l$; the integer problem adds $x_{li} \in \{0,1\}$, so that $r_l = \sum_i x_{li}$.

\paragraph{Water-filling allocation.}
Sort all directions by $v_{li}/c_l$ in decreasing order, ties within a layer by increasing $i$; as $v_{li}$ is non-increasing in $i$ and $c_l$ is constant within a layer, every initial segment is prefix-feasible. Fill the budget in this order, taking a fraction of the first direction that does not fit; call the result $x^\ast$.

\begin{proof}[Proof of Proposition~\ref{prop:waterfill}]
If $B \ge C(\mathbf 1)$ then $x^\ast = \mathbf 1$, which maximizes a non-negative objective. Otherwise $C(x^\ast) = B$; let $\lambda \ge 0$ be the ratio $v_{li}/c_l$ of the boundary direction, so that $x^\ast_{li} = 1$ whenever $v_{li}/c_l > \lambda$ and $x^\ast_{li} = 0$ whenever $v_{li}/c_l < \lambda$. For any feasible $x$ the term $(v_{li} - \lambda c_l)(x_{li} - x^\ast_{li})$ is non-positive for every $(l,i)$: above the threshold the first factor is positive and $x_{li} \le 1$, below it the first factor is negative and $x_{li} \ge 0$, and on it the first factor vanishes. Summing,
\[
E(x) - E(x^\ast)
= \sum_{l,i} (v_{li} - \lambda c_l)(x_{li} - x^\ast_{li}) + \lambda\,[\,C(x) - C(x^\ast)\,]
\le \lambda\,[\,C(x) - B\,] \le 0 ,
\]
so $x^\ast$ is optimal for the relaxation. If $x^\ast$ is integral it is integer-feasible and attains the relaxation bound, hence integer-optimal; this covers every budget ending exactly at a prefix of the ratio order. If all layers share one cost $c$, every integer-feasible allocation keeps at most $k = \min(M, \lfloor B/c \rfloor)$ of the $M$ directions, so keeping the $k$ largest energies (with prefix-preserving ties) is integer-optimal whether or not $B/c$ is an integer.
\end{proof}

\begin{corollary}[Penalized objective]
\label{cor:penalized}
For any $\lambda \ge 0$, the prefix-feasible binary selection $x_{li} = \mathbf 1[\,v_{li} > \lambda c_l\,]$ (with ties resolved either way) maximizes $E(x) - \lambda\, C(x)$ over all prefix-feasible binary $x$.
\end{corollary}
\begin{proof}
The objective separates as $\sum_{l,i} (v_{li} - \lambda c_l)\, x_{li}$; keeping exactly the positive coefficients is optimal term by term and prefix-feasible, since the coefficients are non-increasing in $i$ within each layer.
\end{proof}

\paragraph{What is and is not covered.}
Corollary~\ref{cor:penalized} is exact for heterogeneous costs under a storage \emph{penalty} and yields a hard-budget optimum only in the integral case of the proposition; otherwise the whole-direction greedy allocator can fall short of the integer optimum. Against an exact dynamic program over all $391$ parameter tensors and $222{,}805$ directions of each of the $30$ SD-v1.5 subjects, its mean relative energy shortfall is $2.5{\times}10^{-5}$ at $5$\,MiB and $9.0{\times}10^{-6}$ at $13$\,MiB (worst subjects $7.5{\times}10^{-5}$ and $2.6{\times}10^{-5}$), and the gap to the relaxation bound shrinks from $3.8{\times}10^{-5}$ at $5$\,MiB to $6.2{\times}10^{-7}$ at $151$\,MiB. The fractional step is used only to compute that bound; the codec stores whole ranks.

\end{document}